\documentclass[10pt,twocolumn,letterpaper]{article}

\usepackage[pagenumbers]{wacv} 

\definecolor{wacvblue}{rgb}{0.21,0.49,0.74}
\usepackage[pagebackref,breaklinks,colorlinks,allcolors=wacvblue]{hyperref}
\usepackage{placeins}
\usepackage{pifont}
\usepackage{makecell}
\usepackage{siunitx}
\usepackage{multirow}
\usepackage{graphicx}
\usepackage{subcaption}
\usepackage{xcolor}

\newcommand{\bfqty}[2]{\text{\bfseries\SI{#1}{#2}}}
\DeclareSIUnit{\px}{px}

\def\wacvPaperID{2159} 
\def\confName{WACV}
\def\confYear{2026}

\title{Uplifting Table Tennis: A Robust, Real-World Application for 3D Trajectory and Spin Estimation}

\author{Daniel Kienzle\\
{ \normalsize University of Augsburg }\\
{\tt\small daniel.kienzle@uni-a.de}
\and
Katja Ludwig\\
{ \normalsize University of Augsburg }\\
{\tt\small katja.ludwig@uni-a.de}
\and
Julian Lorenz\\
{ \normalsize University of Augsburg }\\
{\tt\small julian.lorenz@uni-a.de}
\and
Shin'Ichi Satoh\\
{ \normalsize National Institute of Informatics \& University of Tokyo} \\
{\tt\small satoh@nii.ac.jp}
\and
Rainer Lienhart\\
{ \normalsize University of Augsburg }\\
{\tt\small rainer.lienhart@uni-a.de}
}

\begin{document}
\maketitle
\begin{abstract}
Obtaining the precise 3D motion of a table tennis ball from standard monocular videos is a challenging problem, as existing methods trained on synthetic data struggle to generalize to the noisy, imperfect ball and table detections of the real world.
This is primarily due to the inherent lack of 3D ground truth trajectories and spin annotations for real-world video.
To overcome this, we propose a novel two-stage pipeline that divides the problem into a front-end perception task and a back-end 2D-to-3D uplifting task.
This separation allows us to train the front-end components with abundant 2D supervision from our newly created TTHQ dataset, while the back-end uplifting network is trained exclusively on physically-correct synthetic data.
We specifically re-engineer the uplifting model to be robust to common real-world artifacts, such as missing detections and varying frame rates.
By integrating a ball detector and a table keypoint detector, our approach transforms a proof-of-concept uplifting method into a practical, robust, and high-performing end-to-end application for 3D table tennis trajectory and spin analysis.
\end{abstract}

\section{Introduction}
\label{sec:intro}
Table tennis is a dynamic sport demanding exceptional precision, speed, and strategic thinking.
For athletes, coaches, and sports scientists, understanding the intricate 3D trajectory and spin of the ball is paramount for in-depth performance analysis and technique refinement.
Such detailed insights, however, are notoriously difficult to obtain from conventional broadcast or monocular video footage, which only provides 2D observations.
The rapid motion of the ball, coupled with occlusions, varying lighting conditions, and diverse camera angles, poses significant challenges for accurate 3D reconstruction.
\begin{figure}[t]
    \centering
    \includegraphics[width=0.97\linewidth]{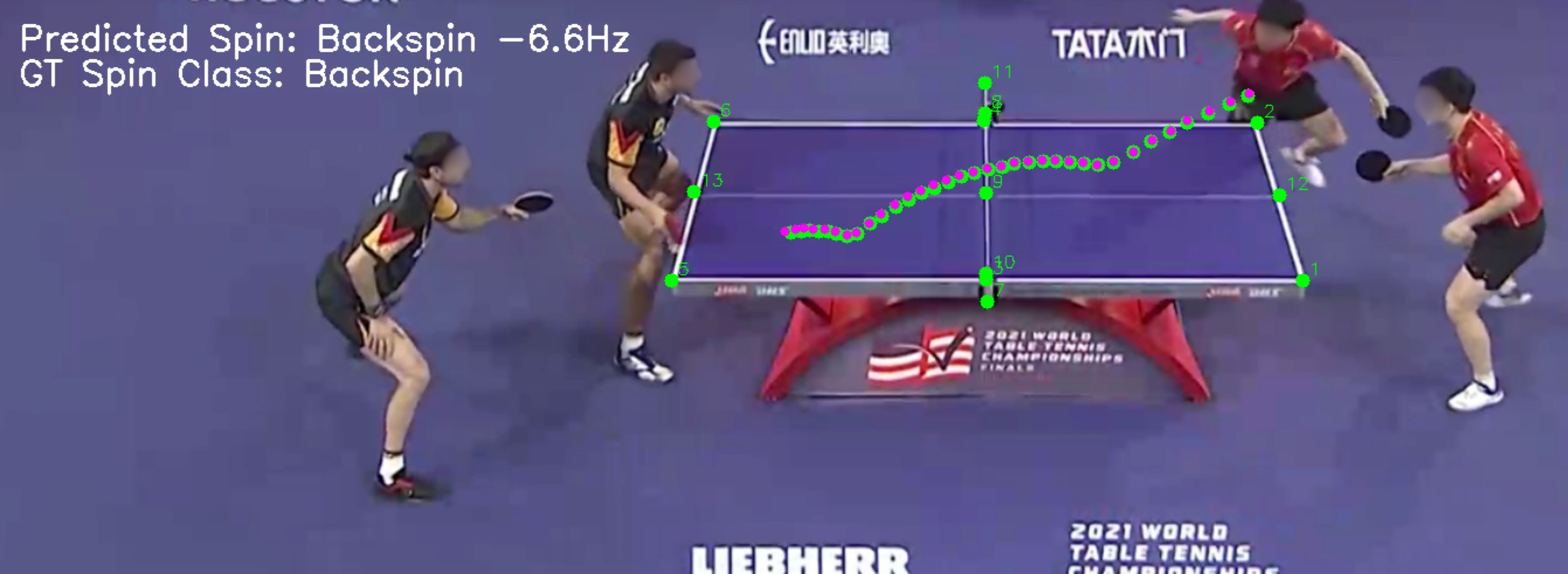}
    \vspace{-0.1cm}
    \caption{Qualitative example prediction of the full pipeline for a serve trajectory.
    The \textcolor[rgb]{0.0,0.99,0.0}{green} dots represent the front-end detections for 2D ball positions and table keypoints.
    The \textcolor[rgb]{0.99,0.0,0.99}{magenta} dots represent the predicted 3D ball trajectory from the back-end.
    }
    \label{fig:3D_reconstruction_pipeline}
\vspace{-0.40cm}
\end{figure} 
Prior research has demonstrated the feasibility of training neural networks on synthetically generated data to reconstruct 3D trajectories and spin \cite{synthnet,tennis3Dtrajectory,spinAnd3DTrajectory}.
However, a significant gap remains in transitioning these models from clean, synthetic inputs to the noisy, sparse, and imperfect detections found in real-world videos.
This discrepancy severely limits their practical deployment.
This paper addresses these limitations by presenting a comprehensive pipeline designed to robustly infer the 3D trajectory and initial spin of a table tennis ball directly from monocular video footage.
To the best of our knowledge, this work constitutes the first learning-based application of a complete pipeline for 3D table tennis analysis. \\[0.33ex]
Our solution is built on a novel two-stage framework that solves the fundamental problem of missing 3D ground truth in real-world broadcast videos.
We achieve this by dividing the problem into a front-end perception stage and a back-end uplifting stage, which allows us to train each component with different, readily available supervision.
We also introduce a new dataset, essential for training our models. \\[0.33ex]
Our main contributions are:
\begin{itemize}
\vspace{-0.02cm}
\item We develop a state-of-the-art \textbf{2D ball detector} leveraging the efficient Segformer++ architecture \cite{segformer++}, specifically optimized for processing high-resolution images.
\item We introduce a novel \textbf{2D table keypoint detector}, enabling precise localization of the table boundaries within diverse video frames.
These keypoints provide essential contextual information for the 2D-to-3D uplifting model.
\item We present a \textbf{2D-to-3D uplifting model} that takes the detected 2D ball trajectory and table keypoints as input, outputting the 3D trajectory and initial spin of the ball.
Even though the model is solely trained on physically-correct synthetic data, it achieves zero-shot generalization to real-world scenarios.
We specifically adapt this model to robustly handle real-world detection noise, missing detections, and varying frame rates, making it compatible with our presented detectors.
\item We introduce the \textbf{TTHQ dataset}, a novel high-quality, high-resolution dataset featuring meticulously annotated 2D ball trajectories, annotated table keypoints, spin information, and comprehensive meta information, all sourced from publicly available YouTube videos.
This dataset is instrumental for training and evaluating our 2D detectors and for benchmarking the integrated pipeline.
\end{itemize}

\section{Related Work}
\label{sec:relwork}
\noindent \textbf{Ball and Table Keypoint Detection} \\[0.33ex]
General object detectors \cite{yolo,fasterrcnn,detr} can be adapted for ball detection, but heatmap-based methods, common in 2D pose estimation \cite{convolutionalposemachines,stackedhourglassnetworks,freelyselectedkeypoints,vitpose}, have become the de facto standard for ball detection.
The TrackNet model family \cite{tracknetv1,tracknetv2,tracknetv3} and the state-of-the-art WASB \cite{wasb}, demonstrate strong performance.
In addition to the ball position, we also want to detect specific table keypoints in the image, which is also sometimes performed for camera calibration in sports analytics \cite{rinkagnostichockey,nobellsjustwhistles,KaliCalib}.
However, existing methods in table tennis \cite{tt3d,lattemv} often lack the precision and comprehensive set of points required for direct integration into our pipeline.
We address these limitations by leveraging the \textbf{Segformer++} architecture \cite{segformer++}.
This modern, transformer-based approach is uniquely suited for our task due to its efficiency in processing \textbf{high-resolution images}, which is crucial for capturing tiny ball details and thin table edges. 
It is trained using a heatmap-based approach, which proves exceptionally effective for precise ball localization.
\\[1ex]
\noindent \textbf{3D Trajectory and Spin Estimation} \\[0.33ex]
Reconstructing 3D trajectories from monocular video is challenging.
While controlled multi-camera setups offer high accuracy via triangulation \cite{triangulation1,triangulation2,triangulation3,triangulation4}, they are impractical for broadcast footage.
Monocular methods often rely on physics-based model fitting to observed 2D trajectories \cite{MonoTrackBadminton,3DRegressionVolleyball,3DRegressionBasketball,tt3d,lattemv} or single-frame estimates using cues like observed ball size or height \cite{BallLocalizationSingleImages,BallLocalizationSingleImages2,BallLocalizationSingleImagesPhysics,TableTennisBallSize}, but these approaches are susceptible to errors from inaccurate identification of key events, explicit camera calibration or insufficient video quality.
The use of deep learning has shown great promise in overcoming these issues by directly predicting 3D trajectories \cite{synthnet,tennis3Dtrajectory}.
Especially the work of \textit{Kienzle et al.} \cite{spinAnd3DTrajectory} is of great importance for this paper, demonstrating that a network trained only on synthetic data can achieve zero-shot generalization to real-world footage.
However, this is only a proof-of-concept, not considering the front-end detection task and, thus, lacking the necessary robustness for a practical application.
Their approach is not designed to handle \textbf{real-world imperfections} such as imperfect 2D detector outputs, missing detections due to occlusions, and varying video frame rates.
Our work directly addresses this gap.
We \textbf{re-engineer and adapt} this foundational uplifting model to function with imperfect real-world inputs and integrate it into a complete application pipeline.
We also expand the synthetic training dataset to include a wider range of match situations, such as serves and faults, which is essential for building a truly useful application. \\[0.33ex]
Estimating ball spin is equally vital for player analysis but remains a significant challenge due to its indirect observability.
Existing solutions often rely on specialized hardware such as event cameras \cite{TableTennisSpinEventCamera,TableTennisSpinEventCamera2} or high-speed cameras \cite{SpinDOE}.
To our knowledge, \cite{spinAnd3DTrajectory} introduced the first straightforward learning-based method to estimate ball spin from standard monocular video.
Building upon this, we further adapt and improve this spin estimation capability, integrating it robustly within our pipeline. \\[1ex]
\noindent \textbf{Datasets} \\[0.33ex]
Robust applications are highly dependent on high-quality datasets.
Existing table tennis datasets like TTST \cite{spinAnd3DTrajectory} and Blurball \cite{blurball} have been instrumental in previous research, but they lack the necessary scale, resolution, or annotation diversity to train and validate a complete, robust pipeline.
To solve this, we introduce the \textbf{TTHQ dataset}.
This is the first comprehensive, high-resolution dataset sourced from real-world broadcast footage that provides all the necessary annotations - 2D ball positions, table keypoints, and spin labels - in sufficient quantity to train and evaluate an end-to-end pipeline.
This new dataset is a crucial contribution that will enable future research in this domain.

\section{Method}
\label{sec:method}
\begin{figure*}
\centering
\includegraphics[width=0.84\textwidth]{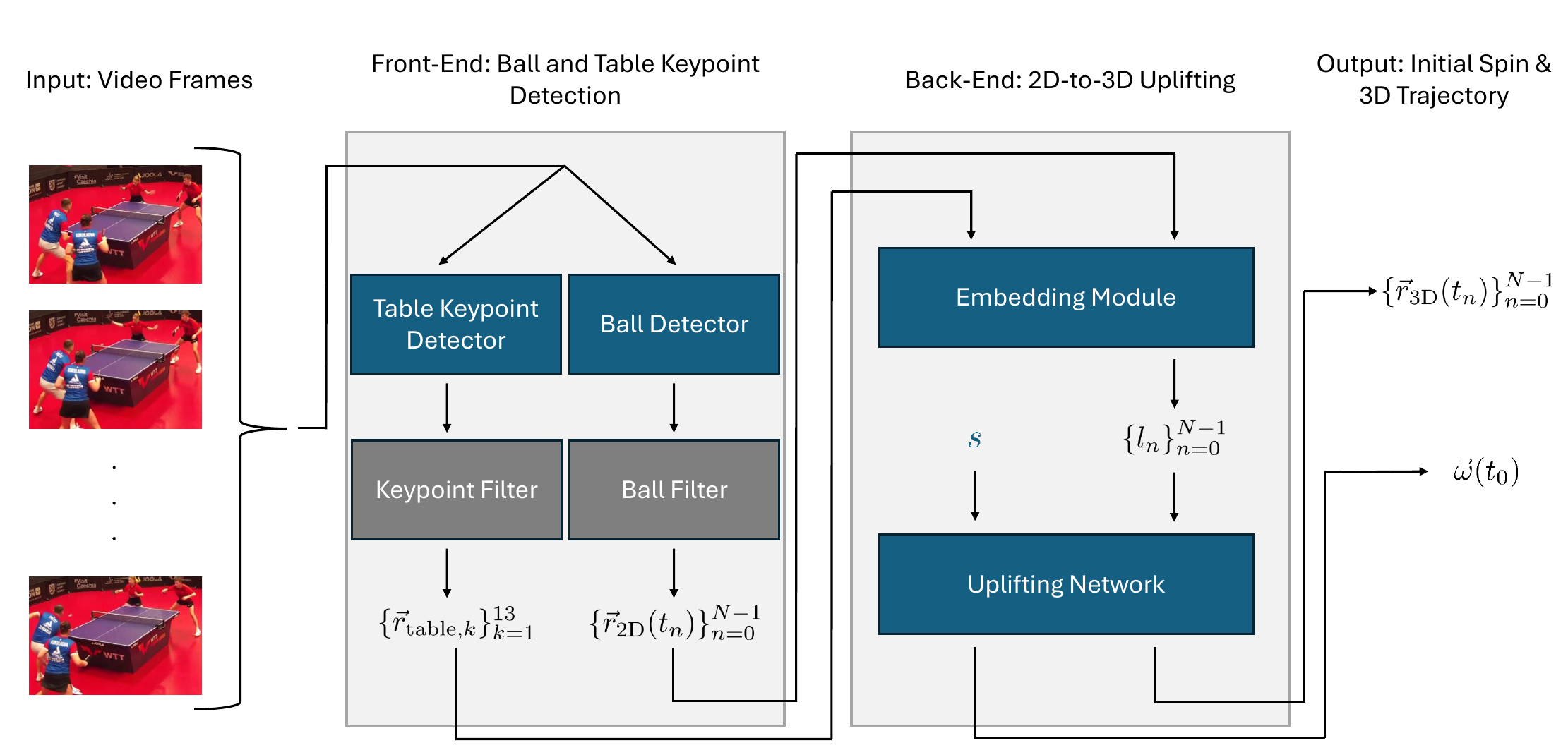}
\vspace{-0.2cm}
\caption{Overview of the proposed pipeline. In the \textbf{front-end stage}, we detect the 2D ball position and localize the 13 table keypoints in each frame $n$ at time $t_n$. After robust filtering, we obtain a clean 2D ball trajectory $\{\vec{r}_\text{2D}(t_n)\}_{n=0}^{N-1}$ with $N$ being the number of frames in the trajectory. As we assume a static camera, we obtain a single, time-independent set of table keypoints $\{\vec{r}_{\text{table,}k}\}_{k=1}^{13}$ after filtering. In the \textbf{back-end stage}, the coordinates are embedded into a location token $l_n$ for each timestep $t_n$, a learnable spin token $s$ is prepended, and the sequence is then processed by the uplifting network to predict the 3D trajectory $\{\vec{r}_\text{3D}(t_n)\}_{n=0}^{N-1}$ and the initial spin $\vec{\omega}(t_0)$. The \textcolor[rgb]{0.082,0.373,0.510}{blue} color represents modules with learnable parameters.
}
\label{fig:overview}
\vspace{-0.5cm}
\end{figure*}

\subsection{General Overview}
\label{sec:general_overview}
The lack of 3D supervision in real-world broadcast videos presents a major challenge for learning the ball's 3D trajectory and spin.
We overcome this core problem by proposing a complete, end-to-end pipeline that divides the task into two distinct stages, each solvable with available supervision.
An overview of our method is given in Figure~\ref{fig:overview}. \\[0.33ex]
The process begins with the \textbf{front-end stage}, which handles the perception and data extraction.
Our robust 2D ball detection module is applied to each frame $n$ at time $t_n$, identifying the ball's 2D position $\vec{r}_\text{2D}(t_n)$.
This process yields a sequence of 2D ball detections given by $\{\vec{r}_\text{2D}(t_n)\}_{n=0}^{N-1}$ with $N$ being the number of frames in the trajectory.
Concurrently, the 2D table keypoint detection module localizes the 13 characteristic table points in each frame.
To enhance accuracy and stability against detection noise, we apply effective filtering techniques to remove erroneous detections.
Moreover, as the camera in table tennis is usually static, we merge the table keypoint detections over time to obtain a single, high-quality set of table keypoints $\{\vec{r}_{\text{table,}k}\}_{k=1}^{13}$ for the entire trajectory. \\[0.33ex]
In the subsequent \textbf{back-end stage}, this processed 2D data is fed into our 2D-to-3D uplifting network.
The network takes the 2D ball trajectory $\{\vec{r}_\text{2D}(t_n)\}_{n=0}^{N-1}$ and the table keypoints as input to predict the initial spin $\vec{\omega}(t_0) \in \mathcal{R}^3$ and the full 3D trajectory $\{\vec{r}_\text{3D}(t_n)\}_{n=0}^{N-1}$ in metric world coordinates.
This modular approach is central to our solution: we train the front-end components on our newly created TTHQ dataset, which provides extensive manually-annotated 2D supervision, while the back-end uplifting network is trained solely on synthetic data, similar to \cite{spinAnd3DTrajectory}.
This deliberate separation into two stages solves the problem of missing 3D ground truth for real-world videos, making our complete system a practical application.

\subsection{Ball Detection}
\label{sec:ball_detection}
Precise 2D ball detection is a crucial prerequisite for accurate 3D trajectory reconstruction.
This task is particularly challenging due to the ball's tiny size, high speed, and susceptibility to motion blur and occlusions.
To address these challenges, we require a model that is both computationally efficient for high-resolution imagery and robust to these visual artifacts.
While existing methods like WASB \cite{wasb} use the HRNet architecture \cite{hrnet}, which is effective but computationally expensive, we instead adopt the more efficient Segformer++ architecture \cite{segformer++}.
This transformer-based model leverages token merging \cite{tome} to reduce computational cost, enabling us to process high-resolution inputs, which are essential to prevent the ball from vanishing due to downsampling. \\[0.33ex]
To provide the network with vital temporal context for challenging cases like motion blur and occlusions, we take three consecutive images of height $H$ and width $W$ as input.
These images are concatenated along the channel dimension to form a 9-channel tensor $I \in \mathbb{R}^{H \times W \times 9}$.
The network then outputs a heatmap where each pixel value indicates the confidence of the ball's presence in the central frame.
We obtain the 2D ball position by first locating the pixel with the highest confidence and then refining its position by fitting a 2D Gaussian to the local neighborhood.
Sequentially applying this network to each triple of input frames yields a complete sequence of 2D ball detections given by $\{\vec{r}_\text{2D}(t_n)\}_{n=0}^{N-1}$. \\[0.33ex]
To ensure a clean trajectory, which is crucial for the subsequent uplifting, we present a robust filtering technique to mitigate the impact of false detections and occlusions.
Even highly performant models sometimes detect sporadic false positives, such as a player's feet or paddle, or yield unreliable predictions during occlusions.
Common temporal filtering techniques are insufficient for this task, as false positives often occur for multiple consecutive frames, e.g. tracking the paddle movement and leading to wrong trajectories.
To address this, we utilize an auxiliary model based on the HRNet architecture.
Since the two architectures tend to make different types of errors, we leverage their disagreement to filter out erroneous predictions.
A detection is only considered valid if both models' predictions are in close agreement.
If the models agree, we keep the Segformer++ prediction due to its high accuracy.
In cases where the models disagree, we simply discard the detection.
This simple yet effective filtering step removes nearly all false positives, providing a highly reliable, albeit potentially incomplete, 2D trajectory that is essential for the stable performance of our back-end uplifting network.
We provide visualizations of the filtering in Figure~\ref{fig:ball_filtering} of the supplementary material.

\subsection{Table Keypoint Detection}
\label{sec:table_keypoint_detection}
\begin{figure}
    \centering
    \includegraphics[width=0.70\linewidth]{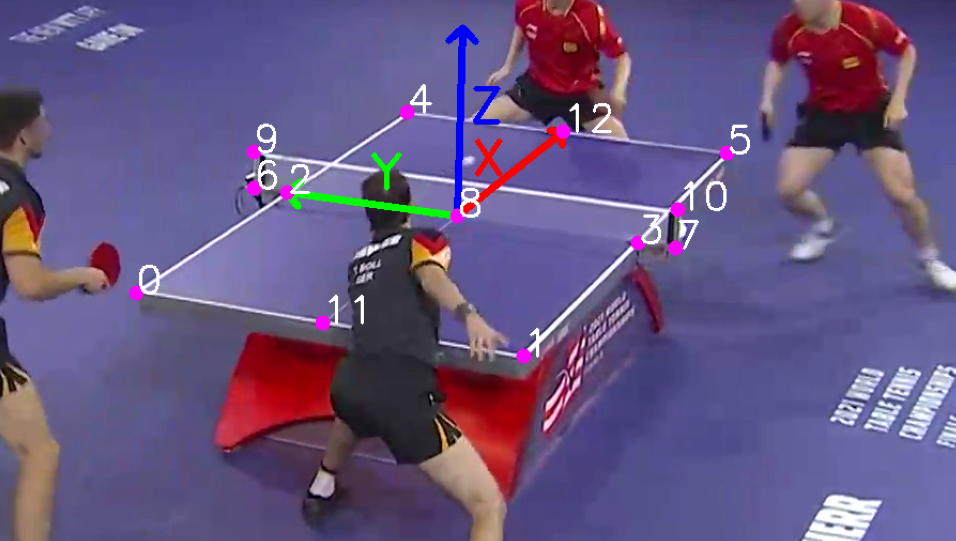}
    \vspace{-0.1cm}
    \caption{Definition of the 13 Table Keypoints and illustration of the world coordinate system axes.}
    \label{fig:keypoint_definition}
\vspace{-0.50cm}
\end{figure}
The table keypoints are a critical input for our uplifting model as they provide a stable spatial reference, enabling the network to understand the camera's perspective and the ball's position relative to the table.
This approach bypasses the need for explicit camera calibration, which is often a source of error and complexity in other methods.
We detect 13 characteristic keypoints, including the four corners, midpoints of each side, and key points along the net.
These keypoints are illustrated in Figure~\ref{fig:keypoint_definition}. \\[0.33ex]
Similar to our ball detection module, we employ the Segformer++ architecture \cite{segformer++} for this task.
However, since the table is static, we use a single frame instead of a sequence of 3 frames as input, which is sufficient to capture the required spatial information.
The network outputs a heatmap for each of the 13 keypoints, from which we extract their 2D positions using the same Gaussian-fitting approach as for ball detection.
Since the different table corners and edges can appear visually similar, the global context of the entire table is essential for accurate keypoint identification, thus, the global receptive field of the Segformer++ architecture is particularly beneficial for this task. \\[0.33ex]
Despite the table's characteristic appearance, detecting its keypoints is challenging due to frequent occlusions by players, paddles, or the ball itself.
We assume a static camera setup during each trajectory, which is standard in table tennis broadcasts, allowing us to aggregate keypoint detections over time to obtain a highly reliable set of keypoints.
We implement a two-step filtering process.
First, we use a cross-architectural filtering technique similar to the one described in Section~\ref{sec:ball_detection}.
We train an auxiliary model based on the HRNet architecture, and compare its predictions with the Segformer++ model.
Since these models have different failure modes, detections are only considered valid if the two models are in close agreement.
Second, we implement a temporal clustering approach to aggregate and refine detections over the entire trajectory.
As some keypoints may be occluded for extended periods, frequently for more than 50\% of the trajectory duration, we cannot rely on simple temporal smoothing techniques.
Therefore, we cluster the valid detections over time for each keypoint using DBSCAN \cite{dbscan}.
The center of the largest cluster is taken as the final, robust keypoint position.
If no cluster is found, the keypoint is considered undetectable for the given video.
Consequently, after this robust filtering, we obtain a single time-independent, high-quality set of table keypoints $\{\vec{r}_{\text{table,}k}\}_{k=1}^{13}$ for the trajectory.
An illustration of this filtering is provided in Figure~\ref{fig:table_filtering} of the supplementary material.

\subsection{2D-to-3D Uplifting}
\label{sec:uplifting_network}
\begin{figure}
\centering
\includegraphics[width=0.35\textwidth]{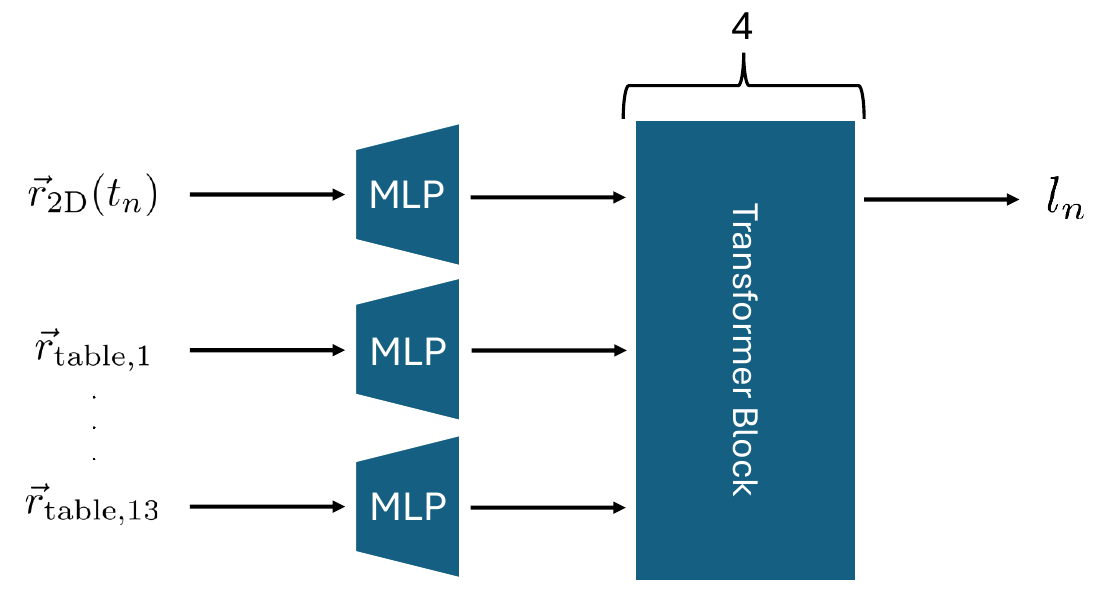}
\vspace{-0.1cm}
\caption{\textbf{Embedding Module}. The detected ball position $\vec{r}_\text{2D}(t_n)$ at time $t_n$ and all visible table keypoints are projected into a higher dimensional space by a 2-layer MLP and then processed by a 4-block transformer. Finally, only the token corresponding to the ball position is kept as location token $l_n$.}
\label{fig:embedding}
\vspace{-0.3cm}
\end{figure}

\begin{figure}
\centering
\includegraphics[width=0.38\textwidth]{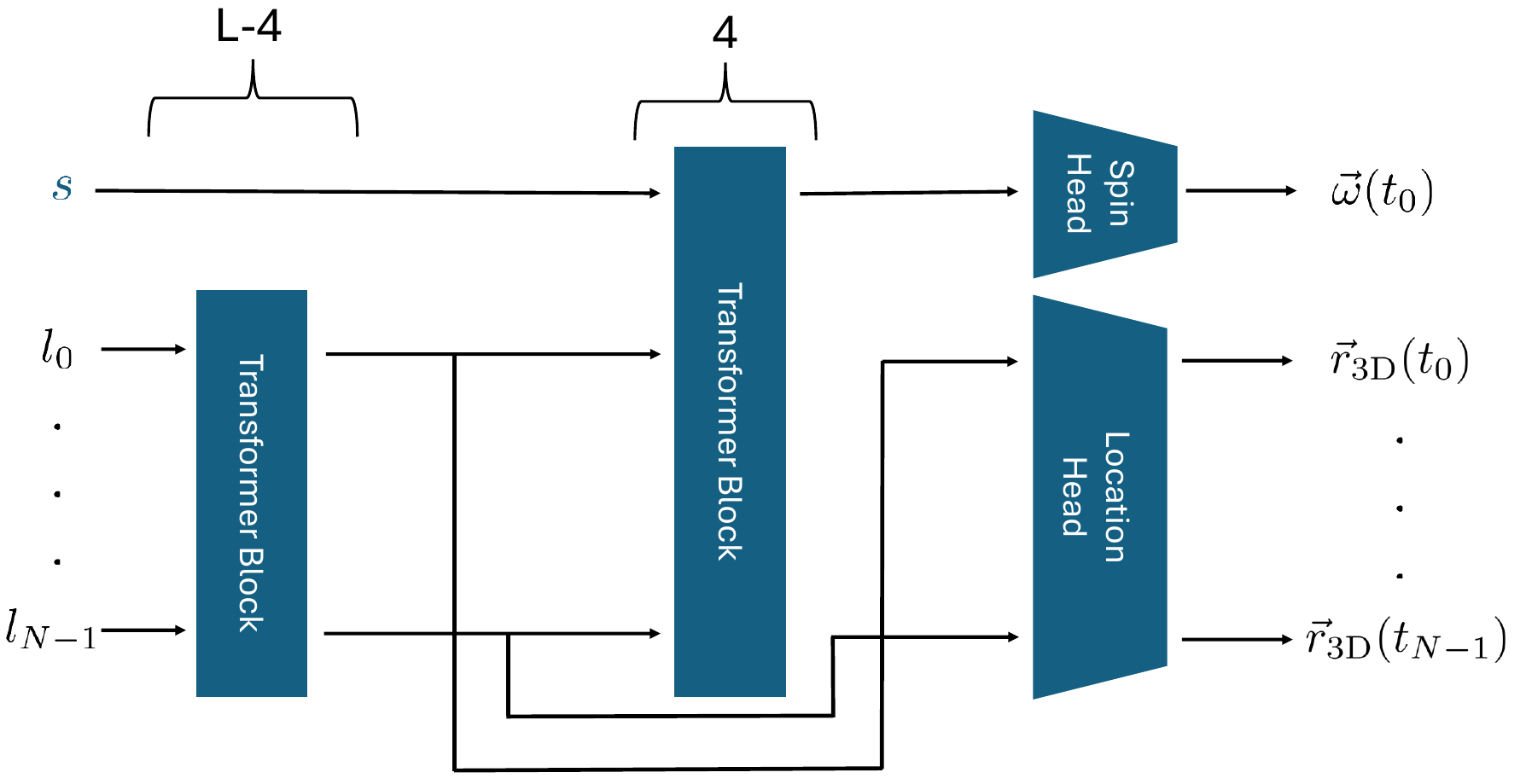}
\caption{\textbf{Uplifting Network}. The input consists of the learnable spin token $s$ and the sequence of location tokens $\{l_0, ..., l_{N-1}\}$ obtained from the embedding module. The first stage consisting of $L-4$ transformer blocks computes the 3D trajectory. The second stage consisting of 4 transformer blocks computes the initial spin. To obtain the final three-dimensional output vectors, we apply a small 3-layer MLP as head for both the trajectory and the spin.}
\vspace{-0.2cm}
\label{fig:uplifting_network}
\vspace{-0.40cm}
\end{figure}

The back-end stage of our pipeline, which performs the 2D-to-3D uplifting, is the central component responsible for reconstructing the 3D data from our front-end 2D detections.
It is designed to be trained solely on synthetic data, which provides perfect 3D ground truth for supervision, thus solving the problem of missing real-world 3D labels.
We build upon the method from \textit{Kienzle et al.} \cite{spinAnd3DTrajectory}, which demonstrates that a network trained on synthetic data can generalize to real-world scenarios.
Given a 2D ball trajectory and the table keypoints, the model predicts the 3D trajectory and initial spin of the ball.
Since they do not consider the front-end detection task, their approach handles manually-annotated 2D inputs, which are perfect and complete.
This is a significant limitation for practical applications, as real-world detections are often imperfect and incomplete due to occlusions and detection errors.
Therefore, we make three key architectural modifications to extend this foundation into a robust, practical application. \\[1ex]
\noindent \textbf{Robustness to Missing Detections} \\[0.33ex]
Our front-end filtering can result in missing detections due to occlusions or false positives.
To handle this naturally, we make use of the network's transformer-based architecture.
The attention mechanism of the transformer allows it to process a sequence of inputs of varying lengths, so we can simply remove the filtered-out ball detections and the network can still process the remaining tokens.
However, the time information between consecutive detections is crucial for accurate trajectory and spin estimation.
Therefore, we employ a custom Rotary Positional Embedding (RoPE) \cite{rope} which encodes the exact time stamps of each detection, as described in more detail below.
Consequently, the network can naturally understand the time difference between consecutive detections, even when some are missing.
To bridge the gap between training on synthetic data and applying it to real-world scenarios, we simulate missing ball detections during training, increasing the network's robustness against these common failure cases. \\[0.33ex]
Similarly, we encode the table keypoints using a custom transformer-based embedding module, which allows us to handle varying numbers of visible keypoints, e.g. due to occlusions.
We also randomly drop keypoints during training to improve robustness to missing keypoint detections. \\[1ex]
\noindent \textbf{Handling of Varying Frame Rates} \\[0.33ex]
Unlike prior work, our system must handle videos with a wide range of frame rates.
To enable the network to understand the time difference between consecutive detections, we use a custom Rotary Positional Embedding (RoPE) \cite{rope}.
Instead of applying a fixed rotation for each position in the input sequence, we rotate each token proportionally to its exact time stamp $t_n$.
This naturally encodes the time information into the attention mechanism.
Further details are provided in Section \ref{sec:architectures} of the supplementary material.
We randomly sample frame rates during training to improve robustness at test time. \\[1ex]
\noindent \textbf{Network Architecture and Data Flow} \\[0.33ex]
Our back-end pipeline begins with an embedding module consisting of $4$ transformer blocks as illustrated in Figure~\ref{fig:embedding}.
We designed this module to process the embedded 2D ball position $\vec{r}_\text{2D}(t_n)$ at time $t_n$ and all visible table keypoints as input.
The transformer blocks distill the context from the keypoints into a single \textbf{location token} $l_n \in \mathcal{R}^d$ for each ball detection.
By calculating an embedding for each detection and prepending a learnable \textbf{spin token} $s \in \mathcal{R}^d$ to the sequence of location tokens $\{l_0, ..., l_{N-1}\}$, we obtain the final input sequence for the uplifting network. \\[0.33ex]
The uplifting network itself is a two-stage transformer as shown in Figure~\ref{fig:uplifting_network}.
First, a 2D-to-3D uplifting is performed by $L-4$ transformer blocks, followed by a simple MLP head to compute the 3D trajectory $\{\vec{r}_\text{3D}(t_n)\}_{n=0}^{N-1}$.
Then, the motion information is distilled into the spin token by $4$ additional transformer blocks, followed by another MLP head to predict the initial spin vector $\vec{\omega}(t_0)$.
This modular design, together with our targeted architectural adaptations, allows our back-end to be exclusively trained on synthetic data and still generalize to real-world footage, effectively solving the challenge of unavailable 3D supervisions.

\section{Data and Metrics}
\label{sec:dataset}
\textbf{TTHQ Dataset} \\[0.33ex]
To train and evaluate our method, we introduce the TTHQ (Table Tennis High Quality) dataset, which is created to address the shortcomings of existing datasets for our specific task.
It consists of high-resolution ($1920\times 1080$) broadcast videos of professional, semi-professional, and amateur table tennis matches sourced from YouTube, covering a wide range of conditions, including different camera angles, lighting situations, and player skill levels.
The TTHQ dataset includes $14$ videos of individual matches and $5$ highlight videos, where a highlight video is a compilation of selected scenes from multiple matches, increasing the diversity of scenarios significantly.
We provide precise 2D annotations for both ball and table keypoint detection modules.
The 2D ball position is manually annotated in $9\,092$ frames, ensuring a diverse set of scenarios with varying ball speeds and motion blur levels.
We also annotate the 13 2D table keypoints in $257$ frames, placing special emphasis on highlight videos and scenes with varying levels of occlusion to ensure the network can generalize despite the relatively small number of annotations. \\[0.33ex]
For comprehensive evaluation, we designate $3$ of the videos for validation and testing.
For these videos, we also provide binary spin class labels (topspin or backspin) for $57$ trajectories, which we annotated similarly to \cite{spinAnd3DTrajectory}.
This allows us to evaluate the full pipeline's spin prediction accuracy in a realistic setting. \\[1ex]
\textbf{Synthetic Data} \\[0.33ex]
To train our uplifting network, we generate a large synthetic dataset using the MuJoCo physics engine \cite{mujoco}.
We extend the original training data from \cite{spinAnd3DTrajectory} by including a much wider variety of trajectories.
This includes trajectories from both sides, along with serves and various fault types.
We also vary the ball's initial position, velocity, and spin more extensively, and we simulate a wider range of camera parameters to cover more diverse real-world broadcast conditions.
Overall, we generate $140\,000$ synthetic trajectories, with over $4.7$ million individual 2D ball positions. \\[1ex]
\textbf{Existing Datasets} \\[0.33ex]
The Blurball dataset \cite{blurball} consists of $25$ videos and provides $50\,000$ annotated frames with 2D ball positions.
While the videos are of lower resolution ($1280\times720$) and quality, we use this dataset for pre-training our ball detection module to bootstrap its learning. \\
The TTST dataset \cite{spinAnd3DTrajectory} provides $50$ trajectories.
The 2D ball position is annotated in $1197$ frames and the set of 13 table keypoints is annotated once per trajectory.
The trajectories are also annotated with a binary spin label (topspin or backspin).
We use $16$ trajectories for validating the uplifting network.
The remaining $34$ trajectories are used to test the uplifting network's performance in a controlled setting. \\[1ex]
\textbf{Metrics} \\[0.33ex]
For our front-end modules, we use the \textbf{\textit{ACC}@$X$px} metric, which measures the percentage of correct detections within a radius of $X$ pixels around the ground truth position. \\[0.33ex]
Given that no 3D ground truth data exists for realistic broadcast videos, a direct evaluation of the back-end's 3D predictions is not possible.
We overcome this challenge by employing a comprehensive set of metrics that indirectly, but robustly, assess our model's performance.
We first evaluate the quality of the predicted 3D trajectory by projecting it back into the image plane.
The 2D Reprojection Error (2DRE) is then calculated as the average distance between the reprojected 3D positions and the ground truth 2D positions for each trajectory:
\begin{equation}
\resizebox{0.70\linewidth}{!}{$
\text{2DRE} = \frac{1}{N} \sum\limits_{n=0}^{N-1} \left\| \mathcal{P} \left( \vec{r}_\text{3D}(t_n) \right) - \hat{\vec{r}}_\text{2D}(t_n) \right\|_2
$}
\end{equation}
where $\mathcal{P} \in \mathcal{R}^{2 \times 3}$ is the camera projection matrix, $\vec{r}_\text{3D}(t_n)$ is the predicted 3D position, $\hat{\vec{r}}_\text{2D}(t_n)$ is the ground truth 2D position, and $N$ is the number of frames in the trajectory.
The final reported score is the \textbf{mean 2DRE (m2DRE)} computed over all trajectories. \\[0.33ex]
Finally, we evaluate the predicted initial spin using the binary spin annotations in the dataset.
Since it is not possible for humans to visually estimate the full 3D spin vector or reliably estimate the strength of the spin, human annotations are inherently limited to the binary classification of topspin vs. backspin.
While our network is trained on synthetic data to predict the full continuous 3D spin vector $\vec{\omega}_\text{3D}(t_0) \in \mathcal{R}^3$, we can only evaluate its performance on real-world data by classifying the spin as topspin if the $y$-component of the predicted vector is positive, and as backspin otherwise.
We perform this classification using a hard threshold, providing a direct mapping from our continuous prediction to the available ground truth.
The performance on this binary task provides compelling evidence that the network has successfully learned the underlying full 3D spin vector.
We calculate the \textbf{Accuracy} (\textit{ACC}) and the \textbf{macro F1-Score} ($F_1$) to measure the classification performance.

\section{Experiments}
\label{sec:experiments}
In this section, we present a systematic evaluation of our proposed components and the complete pipeline across various datasets.
The structure of this section is designed to validate each of our key architectural choices and demonstrate the real-world efficacy of our full system.
Implementation and training details can be found in Section~\ref{sec:training_details} of the supplementary material.

\subsection{Ball Detection Evaluation}
\label{sec:ball_detection_evaluation}
\begin{table}[h!]
\vspace{-0.2cm}
\centering
\resizebox{\columnwidth}{!}{
\begin{tabular}{{l||c|c|c||c|c|c}}
\toprule
Model & \#Params & Input Res. & FPS $\uparrow$ & \textit{ACC}@2px $\uparrow$ & \textit{ACC}@5px $\uparrow$ & \textit{ACC}@10px $\uparrow$ \\
\midrule
Segformer++ (B0) & $3.7 \cdot 10^6$ & $1920 \times 1088$  & \SI{26}{} & \bfqty{75.0}{\%} & \SI{85.9}{\%} & \SI{90.3}{\%} \\
Segformer++ (B2) & $24.7 \cdot 10^6$ & $1600 \times 896$ & \SI{18}{} & \bfqty{75.0}{\%} & \SI{87.1}{\%} & \bfqty{91.8}{\%} \\
WASB (HRNet Small) & $1.5 \cdot 10^6$ & $1280 \times 704$ & \SI{16}{} & \SI{72.4}{\%} & \bfqty{87.4}{\%} & \SI{91.3}{\%} \\
VitPose (ViT Small) & $25.9 \cdot 10^6$ & $1152 \times 640$ & \SI{19}{} & \SI{38.0}{\%} & \SI{50.3}{\%} & \SI{52.1}{\%} \\
\bottomrule
\end{tabular}
}
\vspace{-0.1cm}
\caption{Comparison of different architectures for \textbf{ball detection} and results on the TTHQ test set.
The input resolution for each model was chosen individually to ensure comparable GPU RAM usage during training and similar inference speeds during testing.
This process allowed us to determine the optimal and maximal input resolution for each architecture.
The inference FPS are calculated on a single V100 GPU with batch size $8$.}
\label{tab:ball_detection_results}
\vspace{-0.2cm}
\end{table}
\noindent Our core hypothesis for the front-end is that \textbf{high-resolution input is paramount for precise ball detection}.
Because a table tennis ball is a small, fast-moving object, any loss of visual information can be detrimental.
We suggest the \textbf{Segformer++} architecture due to its strong performance and efficiency, which allows us to use very high-resolution inputs.
To validate this choice, we conduct a comparative evaluation against two distinct architectures: the current state-of-the-art ball detection model \textbf{WASB} based on the HRNet architecture and the popular \textbf{VitPose} architecture for 2D human pose estimation. 
More details about the chosen model parameters are given in Section \ref{sec:architectures} of the supplementary material. \\[0.33ex]
The results in Table~\ref{tab:ball_detection_results} confirm our hypothesis, as the Segformer++ models outperform the other architectures, especially on the strict \textit{ACC}@2px metric.
This high precision is vital, as it is a prerequisite for accurate 3D trajectory uplifting.
While the Segformer++ (B2) model achieves superior overall performance, the smaller B0 variant provides a strong balance of accuracy and speed.
The CNN-based WASB and transformer-based VitPose models perform worse, which we attribute to their lower effective resolution.
This confirms our core belief: the ability to process high-resolution images is paramount for accurate table tennis ball detection.
Based on these results, we select the \textbf{Segformer++ (B2)} model for our final pipeline due to its superior overall performance.

\subsection{Table Keypoint Detection Evaluation}
\label{sec:table_keypoint_evaluation}
\begin{table}[h!]
\vspace{-0.2cm}
\centering
\resizebox{\columnwidth}{!}{
\begin{tabular}{{l||c|c|c||c|c|c}}
\toprule
Model & \#Params & Input Res. & FPS $\uparrow$ & \textit{ACC}@2px $\uparrow$ & \textit{ACC}@5px $\uparrow$ & \textit{ACC}@10px $\uparrow$ \\
\midrule
Segformer++ (B0) & $3.7 \cdot 10^6$ & $1920 \times 1088$  & \SI{26}{} & \SI{43.2}{\%} & \bfqty{86.8}{\%} & \bfqty{94.4}{\%} \\
Segformer++ (B2) & $24.7 \cdot 10^6$ & $1600 \times 896$ & \SI{19}{} & \bfqty{54.3}{\%} & \SI{85.3}{\%} & \SI{93.0}{\%} \\
WASB (HRNet Small) & $1.5 \cdot 10^6$ & $1280 \times 704$ & \SI{17}{} & \SI{41.1}{\%} & \SI{83.8}{\%} & \SI{89.3}{\%} \\
VitPose (ViT Small) & $25.3 \cdot 10^6$ & $1152 \times 640$ & \SI{19}{} & \SI{30.0}{\%} & \SI{68.5}{\%} & \SI{79.7}{\%} \\
\bottomrule
\end{tabular}
}
\vspace{-0.1cm}
\caption{Comparison of different architectures for \textbf{table keypoint detection} and results on the TTHQ test set.
The input resolution for each model was chosen individually to ensure comparable GPU RAM usage during training and similar inference speeds during testing.
This process allowed us to determine the optimal and maximal input resolution for each architecture.
The inference FPS are calculated on a single V100 GPU with batch size $8$.}
\label{tab:table_keypoint_results}
\vspace{-0.2cm}
\end{table}
\noindent Our second front-end hypothesis is that additionally to the high-resolution capabilities, a \textbf{global receptive field is crucial for distinguishing table keypoints}.
The 13 table keypoints are often visually similar, requiring a model to understand the context within the entire table to accurately identify them.
This makes a transformer-based architecture with a global receptive field particularly beneficial.
We again compare our \textbf{Segformer++} architecture against the CNN-based \textbf{WASB} and the transformer-based \textbf{VitPose} models. \\[0.33ex]
The results, summarized in Table~\ref{tab:table_keypoint_results}, show that the Segformer++ models significantly outperform the other architectures.
This performance gap is more pronounced than in the ball detection task.
This provides strong evidence for our hypothesis, suggesting that while high resolution is vital, the global context provided by the transformer backbone is essential for accurately distinguishing between the keypoints.
Further evidence comes from the improved performance of the VitPose model compared to the ball detection task, even though it still lags behind.
Based on these results, we again select the \textbf{Segformer++ (B2)} model for our final pipeline due to its superior performance on the critical \textit{ACC}@2px metric.

\subsection{Back-End Evaluation}
\label{sec:3d_uplifting_evaluation}
\begin{table}[t]
\centering
\resizebox{0.79\linewidth}{!}{
\begin{tabular}{l||c|c||c|c|c}
\toprule
\multirow{3}{*}{Method} & \multicolumn{2}{c||}{Transforms} & \multicolumn{3}{c}{Metrics} \\
& \makecell{Half\\FPS} & \makecell{Miss.\\Det.} & $\textit{ACC}$ $\uparrow$ & $F_1$ $\uparrow$ & m2DRE $\downarrow$ \\
\midrule
\textit{Kienzle et al.} \cite{spinAnd3DTrajectory} & \multirow{3}{*}{\ding{53}} & \multirow{3}{*}{\ding{53}} & \SI{97.1}{\%}  & \SI{0.970}{}  & \SI{2.98}{\px}  \\
Mixed &  &  & \bfqty{100.0}{\%} & \bfqty{1.000}{} & \bfqty{2.49}{\px}  \\
Ours &  &  & \SI{97.1}{\%} & \SI{0.970}{} & \SI{3.43}{\px} \\
\midrule
\textit{Kienzle et al.} \cite{spinAnd3DTrajectory} & \multirow{3}{*}{\ding{51}} & \multirow{3}{*}{\ding{53}} & \SI{76.5}{\%} & \SI{0.731}{} & \bfqty{2.71}{\px} \\
Mixed &  &  & \SI{79.4}{\%} & \SI{0.770}{} & \SI{3.13}{\px} \\
Ours &  &  & \bfqty{100.0}{\%} & \bfqty{1.000}{} & \SI{3.54}{\px} \\
\midrule
\textit{Kienzle et al.} \cite{spinAnd3DTrajectory} & \multirow{3}{*}{\ding{53}} & \multirow{3}{*}{\ding{51}} & \SI{88.2}{\%} & \SI{0.876}{}  & \SI{24.15}{\px}  \\
Mixed &  &  & \bfqty{97.1}{\%} & \bfqty{0.970}{} & \bfqty{5.45}{\px} \\
Ours &  &  & \bfqty{97.1}{\%} & \bfqty{0.970}{} & \SI{5.56}{\px} \\
\midrule
\textit{Kienzle et al.} \cite{spinAnd3DTrajectory} & \multirow{3}{*}{\ding{51}} & \multirow{3}{*}{\ding{51}} & \SI{67.7}{\%} & \SI{0.598}{} & \SI{23.54}{\px} \\
Mixed &  &  & \SI{70.6}{\%} & \SI{0.646}{} & \SI{5.99}{\px} \\
Ours &  &  & \bfqty{97.1}{\%} & \bfqty{0.970}{} & \bfqty{5.75}{\px} \\
\bottomrule
\end{tabular}
}
\vspace{-0.1cm}
\caption{Ablation study of the back-end on the TTST dataset~\cite{spinAnd3DTrajectory}. We apply two different transformations to the test set to simulate real-world challenges: The Half FPS transformation drops every second frame to simulate a lower frame rate. The Missing Detections transformation randomly removes $10\%$ of the frames to simulate ball detection failures or frame drops. Additionally, each table keypoint has a $10\%$ chance of being removed.
}
\label{tab:backend_ablation}
\vspace{-0.5cm}
\end{table}
We now turn to the back-end, where our core hypothesis is that our proposed architectural modifications enable the network to \textbf{handle varying frame rates and missing detections} while still maintaining high performance.
As proof, we perform an ablation study with three distinct model variants on the TTST dataset, with results presented in Table~\ref{tab:backend_ablation}. \\[0.33ex]
The models we compare are:
\begin{itemize}
    \item The original architecture from \textbf{\textit{Kienzle et al.}} \cite{spinAnd3DTrajectory}, which uses a different embedding module not designed for missing data and a sequence-based positional embedding that does not encode the exact timing of each frame.
    \item The \textbf{mixed architecture}, which uses our proposed transformer-based embedding module but retains the sequence-based positional encoding.
    \item \textbf{Our full architecture}, which combines the robust transformer-based embedding module with the time-proportional RoPE.
\end{itemize} \vspace{0.33ex}
To test our hypothesis, we augment the TTST test set with two transformations: The \textbf{Half FPS} transformation, which drops every second frame to simulate a lower frame rate. 
And the \textbf{Missing Detections} transformation, which randomly removes $10\%$ of the frames to simulate ball detection failures or frame drops. 
Additionally, each table keypoint has a $10\%$ chance of being removed to simulate occlusions or table keypoint detection failures. \\[0.33ex]
The results provide clear evidence for our design choices.
On the original, unaugmented TTST data, all architectures perform well.
However, when faced with the Half FPS transformation, the spin predictions of \textit{Kienzle et al.} degrade significantly.
The mixed architecture also suffers a notable performance drop in spin accuracy.
This proves that the exact timing of each frame is crucial for accurate spin prediction, and that our time-proportional RoPE is effective in encoding this information.
When applying the Missing Detections transformation, the predicted trajectories of \textit{Kienzle et al.} degrade significantly, while the other two architectures maintain high performance.
This clearly shows that our transformer-based embedding module, which can naturally process sequences of varying lengths, is effective at handling missing data. 
Finally, when both transformations are applied, our architecture outperforms the other two in both spin and trajectory metrics, demonstrating its robustness to real-world challenges. \\[0.33ex]
Overall, our proposed architecture performs robustly in all scenarios, yielding high-quality spin and trajectory predictions.
This validates our design choices and confirms that our method is a practical solution that can handle the complexities of real-world videos.

\subsection{Full Pipeline: Combining Front- and Back-End}
\begin{table}[h!]
\vspace{-0.5cm}
\centering
\resizebox{0.99\linewidth}{!}{
\begin{tabular}{l||c|c|c|c}
\toprule
Dataset & Table: $\text{m2DRE}$ $\downarrow$ & Ball: $\text{m2DRE}$ $\downarrow$ & Spin: $\textit{ACC}$ $\uparrow$ & Spin: $F_1$ $\uparrow$ \\
\midrule
TTHQ & \SI{2.72}{} $\pm$ \SI{5.71}{\px} & \SI{12.28}{} $\pm$ \SI{10.84}{\px} & \SI{89.5}{\%} & \SI{0.900}{} \\
TTST & \SI{5.75}{} $\pm$ \SI{10.26}{\px} & \SI{9.41}{} $\pm$ \SI{16.90}{\px} & \SI{97.1}{\%} & \SI{0.974}{} \\
\bottomrule
\end{tabular}
}   
\vspace{-0.1cm}
\caption{Results of the full pipeline on the TTHQ and TTST datasets. The $\pm$ indicates the standard deviation computed over all trajectories.}
\label{tab:full_pipeline_results}
\vspace{-0.5cm}
\end{table}
\noindent The final and most crucial evaluation is to assess our complete, end-to-end pipeline in a real-world setting.
While the individual components have proven their effectiveness, the true test is whether the noise and imperfections of the front-end detections degrade the back-end's performance significantly.
To evaluate this, we apply our full system to 57 TTHQ trajectories with annotated spin labels and additionally calculate results on the 34 TTST test trajectories. \\[0.33ex]
Since the 57 TTHQ test trajectories lack ball and table keypoint annotations, we assess the performance indirectly using reprojection errors.
We first estimate the camera matrices from the detected table keypoints using a robust RANSAC-based \cite{ransac} calibration algorithm \cite{spinAnd3DTrajectory} to ensure that the calibration is not affected by potential outliers in the keypoint detections. 
Due to the fixed table geometry, we utilize the known 3D positions of the table keypoints and reproject them back into the image.
These reprojected keypoints can then be compared to the detected keypoints to compute a mean 2D reprojection error (m2DRE) in pixels.
Similarly, we reproject the predicted 3D ball trajectory back into the image and compare it to the detected 2D ball positions to compute a ball m2DRE. 
Finally, spin classification is measured directly using the annotated labels. \\[0.33ex]
As shown in Table~\ref{tab:full_pipeline_results}, our pipeline achieves strong performance on both datasets.
Both the table and ball reprojection errors are low, indicating that each component of the pipeline works effectively.
Especially the spin classification achieves high accuracy and $F_1$ scores, which confirms that the data from our front-end is of sufficient quality for the back-end to make accurate predictions.
When we compare the spin classification results on the TTST dataset with the perfect-input results from Section \ref{sec:3d_uplifting_evaluation}, we see a similar high performance.
This is a crucial finding, as it proves that using the front-end detections does not significantly degrade the back-end's spin prediction performance, demonstrating the effectiveness of our complete pipeline as a practical, robust solution for real-world table tennis analysis.
\begin{figure}[ht]
\centering
\begin{subfigure}[b]{0.69\linewidth}
\includegraphics[width=\textwidth]{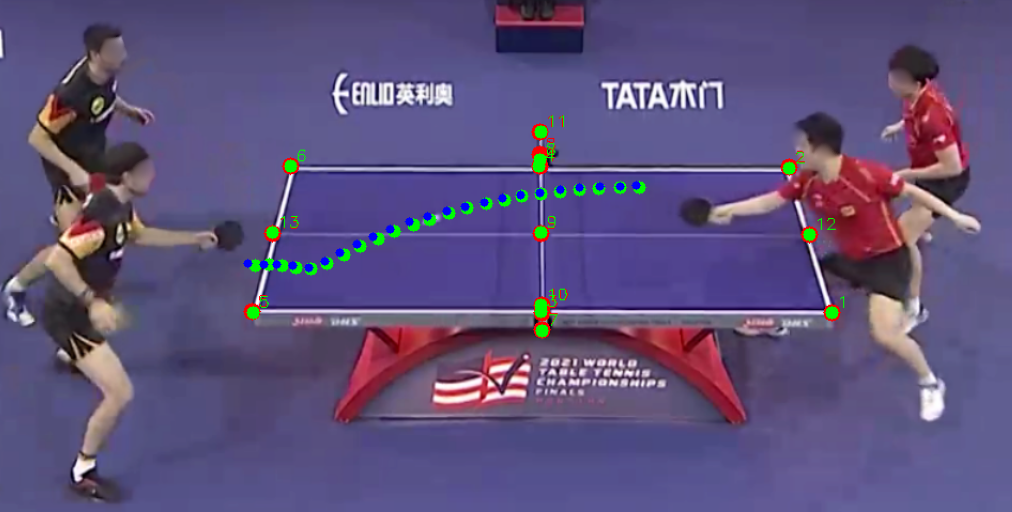} 
\end{subfigure}
\hfill
\begin{subfigure}[b]{0.69\linewidth}
\includegraphics[width=\textwidth]{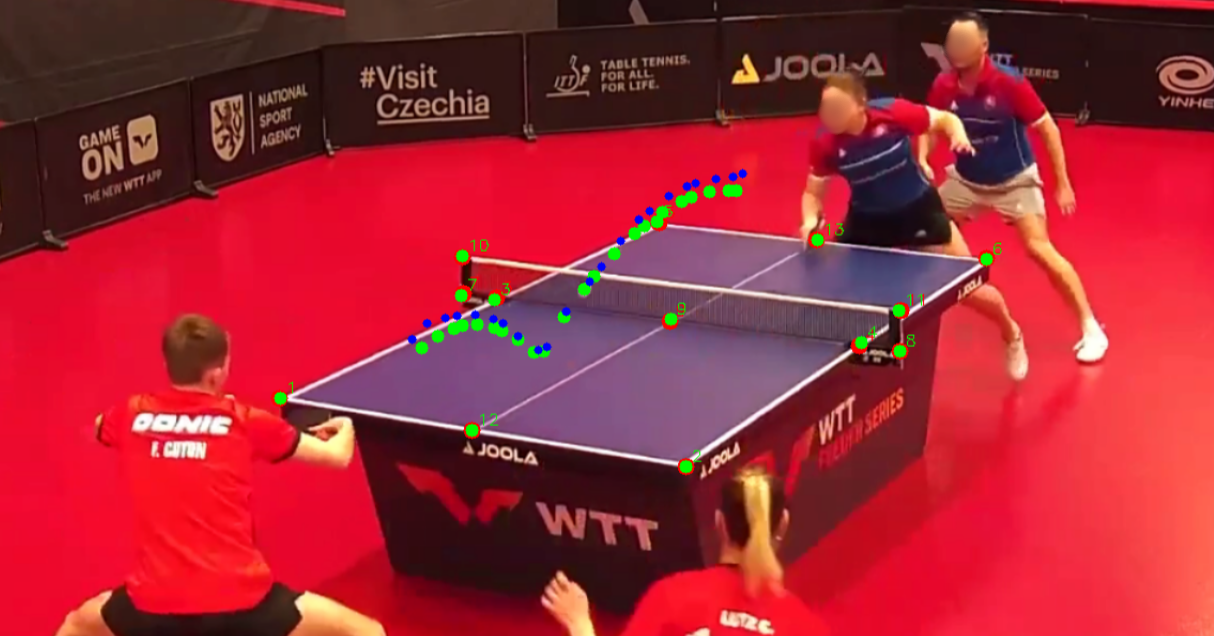} 
\end{subfigure}
\hfill
\begin{subfigure}[b]{0.69\linewidth}
\includegraphics[width=\textwidth]{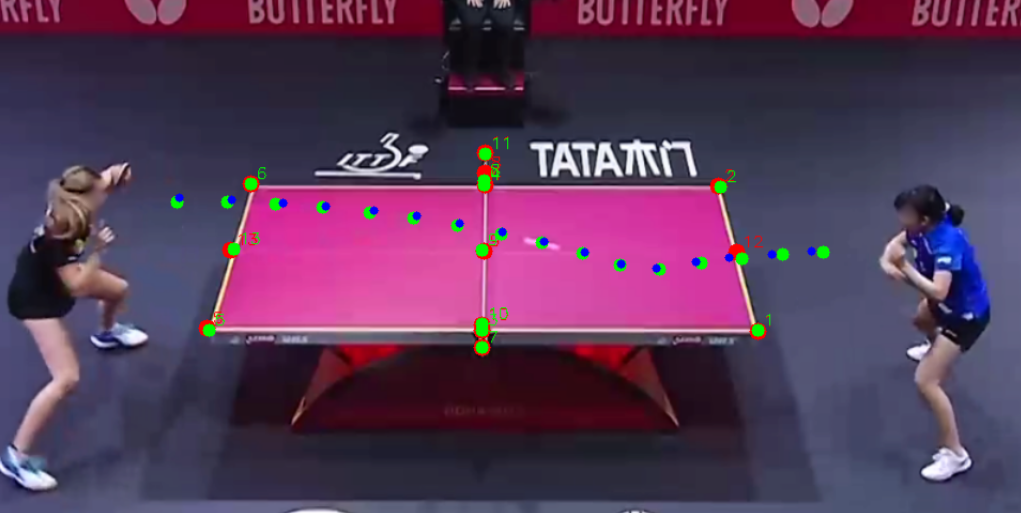} 
\end{subfigure}
\vspace{-0.1cm}
\caption{Qualitative results of the \textbf{full pipeline} on the TTHQ dataset. The \textcolor[rgb]{0.0,0.99,0.0}{green} dots represent the valid detections of the front-end for 2D ball positions and table keypoints. The \textcolor[rgb]{0.99, 0.00, 0.0}{red} dots represent the reprojected table keypoints inferred from the table geometry and camera calibration. The \textcolor[rgb]{0.0, 0.0, 0.99}{blue} dots represent the predicted 3D ball trajectory from the back-end.}
\label{fig:full_results}
\vspace{-0.5cm}
\end{figure}
\section{Conclusion}
\label{sec:conclusion}
In this work, we have presented a comprehensive pipeline for 3D trajectory and spin estimation from monocular video. 
Our approach addresses a novel application domain by providing a solution that can accurately analyze table tennis ball motion from unconstrained, noisy broadcast footage. 
We solve a fundamental challenge in this domain: the lack of 3D ground truth data. \\[0.33ex]
The core of our innovation is a two-stage framework, which provides a clean separation between the 2D perception and 3D uplifting tasks. 
This modularity is a key contribution, as it allows each stage to be trained with different, readily available supervision: 2D human annotations and 3D synthetic data.
Consequently, it eliminates the need for a unified 3D ground truth dataset. 
Our comparative assessment, including extensive ablation studies, validates the superiority of our architectural choices. 
We have explicitly shown that our custom time-proportional positional encoding and robust embedding module are essential for handling real-world artifacts like occlusions and varying frame rates. \\[0.33ex]
Ultimately, our work provides a practical, ready-to-use tool for sports analytics. 
The principles of our two-stage design and our comparative methodology offer a template for similar 3D reconstruction problems in other domains where precise ground truth is a significant barrier.

\newpage

{
    \small
    \bibliographystyle{ieeenat_fullname}
    \bibliography{main}
}

\newpage
\appendix
\clearpage
\maketitlesupplementary

\section{Training Details}
\label{sec:training_details}
The front-end modules are trained with a learning rate of $10^{-3}$ using the Adam optimizer \cite{adam} and a batch size of $4$.
For ball detection, we apply random flipping, rotation, translation, cropping, and color jittering as augmentations.
For table keypoint detection, we do not use flipping, as it would change the semantics of the keypoints.
Instead, we apply an additional random perspective transformation to simulate different camera angles.
Especially for table keypoint detection, we found that augmentations are crucial to achieve generalization ability since the TTHQ dataset is diverse but the number of annotated frames is still limited.
We save the model with the best $\text{\textit{ACC}@}5\text{px}$ metric on the validation set during training and use it for testing.
Similar to 2D human pose estimation, we simply train the models to predict gaussian heatmaps centered at the ground truth ball and keypoint positions. 
The ground truth heatmaps are created at a resolution of $1920\times 1080$, and we use $\sigma = 6\text{px}$ for the Gaussian.
Comparing the predicted and ground truth heatmaps, we use a simple $L_2$ loss.
All evaluations are performed at a resolution of $1920\times 1080$ and the $\text{\textit{ACC}@}X\text{px}$ metric is also computed at this resolution. \\ 
The ball detection backbones are first initialized with ImageNet \cite{imagenet} (Segformer++ and WASB) or MAE \cite{mae} (VitPose) weights, then pre-trained on the Blurball dataset \cite{blurball}, and finally fine-tuned on the TTHQ training set.
The table keypoint detection backbones are also initialized similarly, but only fine-tuned on the TTHQ training set, as the Blurball dataset does not contain sufficiently diverse table keypoint annotations. \\[0.33ex]
For the training of the back-end module, we stick closely to the training procedure of \textit{Kienzle et al.} \cite{spinAnd3DTrajectory}, but we incorporate additional augmentations of the synthetic training data.
We simulate missing detections by randomly dropping $5\%$ of the ball and table keypoint detections during training.
Moreover, we randomly sample the frame rate between \SI{20}{FPS} and \SI{60}{FPS} during training.
A further difference to the \textit{Kienzle et al.} training procedure is that we use our extended synthetic dataset, which contains $140000$ trajectories instead of $50000$ and includes more diverse match scenes like serves.
In contrast to \textit{Kienzle et al.}, we do not validate on the synthetic data, but instead validate on the validation set of the TTST dataset \cite{spinAnd3DTrajectory}.
We choose the final model as compromise between the best $F_1$ score for spin prediction and the best $\text{m2DRE}$ for trajectory prediction.
If multiple epochs achieve the same maximum $F_1$ score, which happens regularly, we choose the one with the best $\text{m2DRE}$ among them.
To allow for a meaningful interpretation of the $\text{m2DRE}$ metric, we evaluate all results at a resolution of $1920\times 1080$, which ensures compatibility with our front-end evaluations.
Our models predict the spin in the global world coordinate system, but for evaluation and extraction of the spin class, we transform the spin into the local ball coordinate system as described in \cite{spinAnd3DTrajectory}.

\begin{figure}[t]
    \centering
    \begin{subfigure}[b]{0.9\linewidth}
        \includegraphics[width=\textwidth]{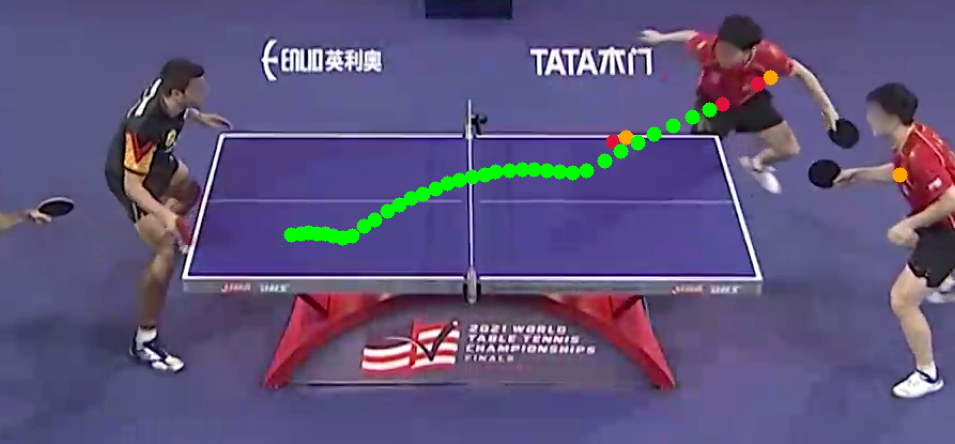} 
    \end{subfigure}
    \hfill 
    \begin{subfigure}[b]{0.9\linewidth}
        \includegraphics[width=\textwidth]{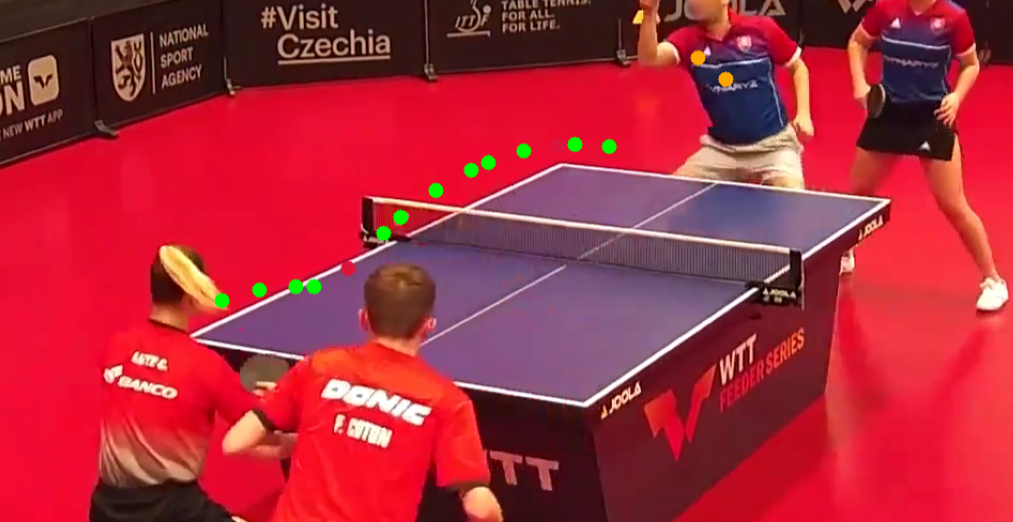} 
    \end{subfigure}
    \hfill 
    \begin{subfigure}[b]{0.9\linewidth}
        \includegraphics[width=\textwidth]{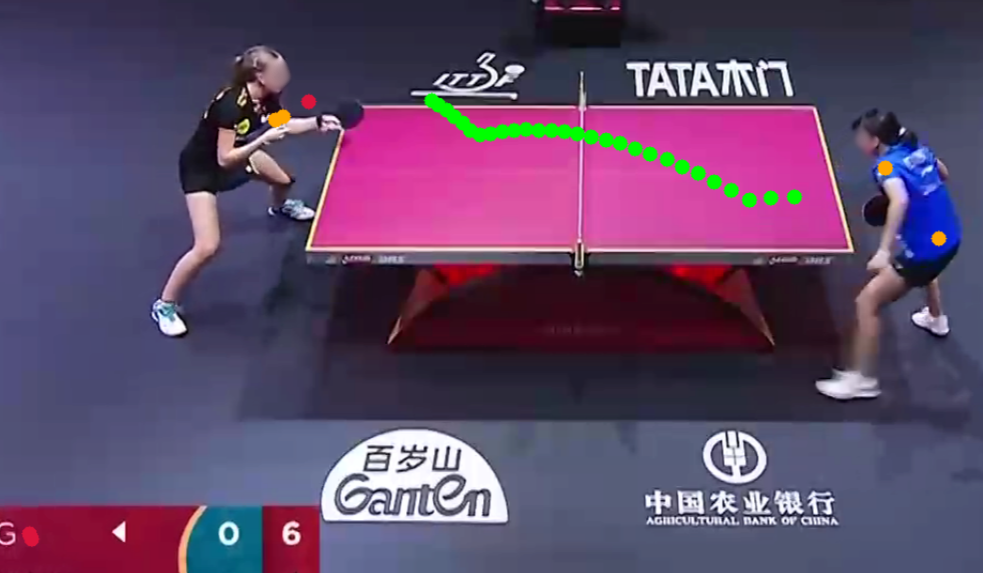} 
    \end{subfigure}
    \caption{Ball detection filtering. The \textcolor[rgb]{0.0,0.99,0.0}{green} dots are the valid detections of the Segformer++ (B2) model after filtering. The \textcolor[rgb]{0.86,0.07,0.24}{crimson} dots are the false positives of the Segformer++ (B2) model that were removed. The \textcolor[rgb]{0.99, 0.65, 0.0}{orange} dots are the predictions of the WASB model for the filtered frames.}
    \label{fig:ball_filtering}
    \vspace{-0.3cm}
\end{figure}

\section{Architectures}
\label{sec:architectures}
\textbf{Front-End Architectures}:
To ensure a fair comparison of front-end architectures, we tuned 2 significant parameters: The input resolution and the model size.
We chose these parameters such that each model can be trained on a single Nvidia V100 GPU with 32 GB of memory. 
Moreover, the models are tuned such that their inference speed is as similar as possible.
We provide the exact parameters in Table~\ref{tab:ball_detection_results} and Table~\ref{tab:table_keypoint_results}. \\[1ex]
\textbf{Back-End Architecture}:
We stick closely to \cite{spinAnd3DTrajectory} when designing the back-end architecture.
We use an embedding size of $d=128$ and $L=16$ transformer layers with $4$ attention heads each.
The Spin Head and Location Head both consist of $3$ fully connected layers with ReLU activations in between. \\[1ex]
\textbf{Positional Encoding}:
Unlike standard transformer-based architectures that rely on a fixed positional encoding based on the token's index, our system must handle inputs with varying time stamps and missing detections.
To address this, we use a custom Rotary Positional Embedding (RoPE) that encodes the exact time information of each detection into the attention mechanism.
The core of our implementation is a modification to the standard RoPE approach, where we replace the discrete position index $n$ with a highly granular, discretized timestamp $p_n$:
\begin{equation}
p_n = \text{round}(t_n / \Delta t)
\end{equation}
with $\Delta t = \SI{2}{\text{ms}}$.
This discretization is so fine-grained that it effectively represents the exact time for all realistic video frame rates.
By replacing $n$ with $p_n$ in the original RoPE formulation \cite{rope}, we obtain the following rotation matrix
\begin{equation}
R = \begin{pmatrix}
\cos(p_n \cdot \theta_m) & -\sin(p_n \cdot \theta_m) \\
\sin(p_n \cdot \theta_m) & \cos(p_n \cdot \theta_m)
\end{pmatrix}
\end{equation}
where $\theta_m = \frac{1}{10000^{2m/d}}$ is the angle, $d$ is the feature dimension and $m$ indexes the feature dimensions.
For a pair of elements from the feature vector $\mathbf{x} = (x_1, ..., x_d)$, the rotated components are computed as:
\begin{equation}
\resizebox{0.9\linewidth}{!}{$
\begin{pmatrix} x_{2m}' \\ x_{2m+1}' \end{pmatrix} = \begin{pmatrix}
\cos(p_n \cdot \theta_m) & -\sin(p_n \cdot \theta_m) \\
\sin(p_n \cdot \theta_m) & \cos(p_n \cdot \theta_m)
\end{pmatrix} \begin{pmatrix} x_{2m} \\ x_{2m+1} \end{pmatrix} \, .
$}
\end{equation}
This time-based encoding allows the attention mechanism to inherently learn the temporal relationships between detections, regardless of their spacing in the input sequence.

\section{Ball \& Table Keypoint Filtering}
\label{sec:filtering}
\begin{figure}[t]
    \centering
    \begin{subfigure}[b]{0.9\linewidth}
        \includegraphics[width=\textwidth]{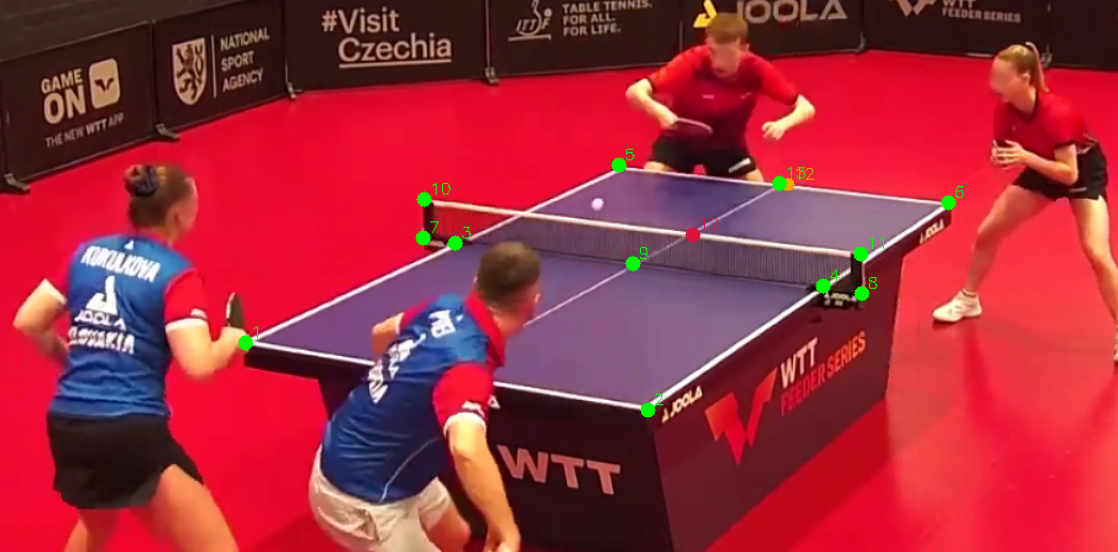} 
    \end{subfigure}
    \hfill 
    \begin{subfigure}[b]{0.9\linewidth}
        \includegraphics[width=\textwidth]{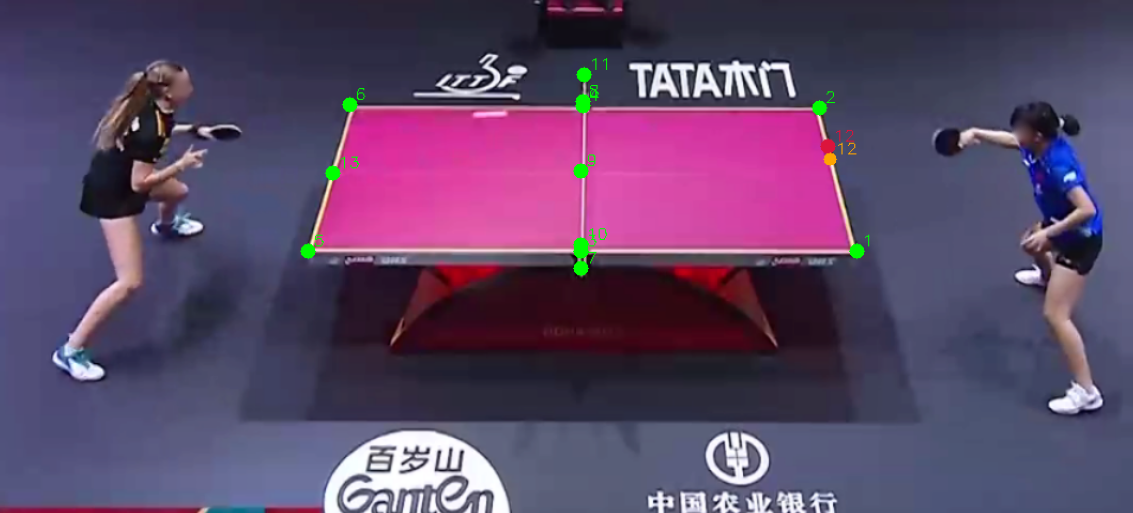} 
    \end{subfigure}
    \vspace{-0.1cm}
    \caption{Table keypoint detection filtering. The \textcolor[rgb]{0.0,0.99,0.0}{green} dots are the valid detections after filtering. The \textcolor[rgb]{0.86,0.07,0.24}{crimson} dots are the false positives of the Segformer++ (B2) model that were removed. The \textcolor[rgb]{0.99, 0.65, 0.0}{orange} dots are the predictions of the WASB model for the filtered frames.}
    \label{fig:table_filtering}
    \vspace{-0.4cm}
\end{figure}

The back-end stage of our pipeline is designed to be robust against missing ball and table keypoint detections.
However, it is relatively sensitive to false positive detections, which is why we apply a filtering step to the raw detections of the front-end.
For ball detections, we compare the predictions of the Segformer++ (B2) model with the predictions of the WASB model and only keep detections with a maximum distance of \SI{20}{px} to a WASB detection.
Because the model architectures are fundamentally different, they usually do not make the same false positive detections, allowing us to filter them out with very high precision.
We illustrate this filtering step in Figure~\ref{fig:ball_filtering} for three different example trajectories in the TTHQ test set. \\
We use this two-model based filtering approach instead of more established filtering methods like Kalman filtering or polynomial fitting, because they fail in a significant failure case: Sometimes, the feet or paddle movements of the players are falsely detected as balls for multiple consecutive frames.
These false positive detections cannot be easily differentiated from real ball trajectories, which is why the established filtering methods fail.
However, our two-model based filtering approach is very robust against this failure case, resulting in a very high precision of the final ball detections. \\[0.33ex]
For table keypoint detections, we apply a similar filtering step. 
We only keep the detections of the Segformer++ (B2) model that are within a distance of \SI{10}{px} to a WASB detection.
Additionally, we utilize the static nature of the table keypoints and apply the DBSCAN clustering algorithm \cite{dbscan} to the remaining detections of each keypoint over the entire trajectory.
The final filtered detections are then the cluster centers of the largest cluster for each keypoint.
This results in an extremely robust filtering of false positive detections.
We illustrate this filtering step in Figure~\ref{fig:table_filtering} for two different example trajectories in the TTHQ test set.

\section{Comparison of Datasets}
\label{sec:datasets}
\noindent We provide a comparison of existing table tennis datasets in Table~\ref{tab:dataset_comparison}. \\[0.33ex]
The \textbf{Blurball dataset} \cite{blurball} has the largest number of ball annotations, and using the provided homographies, pseudo table keypoints can be generated. 
As the dataset is sourced from 24 different videos, the table keypoints lack diversity and are, thus, not suitable for training a robust table keypoint detection model.
However, since ball detection training data needs to capture a wide variety of conditions due to the dynamic nature of the ball, the Blurball dataset is still a valuable resource for training ball detection models.
Due to the low resolution of the videos, we do not use this dataset for fine-tuning, but utilize it for pre-training our ball detection models. \\[0.33ex]
The \textbf{TTST dataset} \cite{spinAnd3DTrajectory} has a high resolution and provides annotations for balls, table keypoints, and spin.
Because the scenes are not very diverse and the number of annotated frames is relatively low, the dataset is not suitable for training robust front-end models.
However, it is still a valuable resource for validating the performance of our pipeline.
Especially valuable are the dense annotations for each trajectory, which allow us to evaluate the performance of the back-end model independently of the front-end. \\[0.33ex]
Our \textbf{TTHQ dataset} provides high resolution videos with a high variety of scenes and numerous annotations for balls, table keypoints, and spin.
This makes it the first dataset that is suitable for training robust front-end models.
Moreover, we can also utilize the dataset for the validation of the spin predictions.
While the annotations are not dense like in TTST, the high variety of scenes and situations makes it an interesting benchmark for evaluating the overall performance of the full pipeline.
\begin{table}[t]
    \centering
    \resizebox{\linewidth}{!}{
    \begin{tabular}{l|c|c|c|c}
        \hline
        \textbf{Dataset} & \textbf{Resolution} & \textbf{\# Balls} & \textbf{\# Table Keypoints} & \textbf{\# Spin Class} \\
        \hline
        TTHQ (ours) & $1920\!\times\!1080$ & 9092 & 257 & 57 \\
        TTST \cite{spinAnd3DTrajectory} & $1920\!\times\!1080$ & 1197 & 50 & 50 \\
        Blurball \cite{blurball} & $1280\!\times\!720$ & 50000 & $50000^*$ & 0 \\
        \bottomrule
    \end{tabular}
    }
    \caption{Comparison of table tennis datasets. 
    We report the video resolution, number of frames with ball annotations, number of frames with table keypoint annotations, and number of trajectories with spin class annotations.
    The * indicates, that table keypoints are not annotated, but can be generated from the provided homography. However, while many frames with table annotations can be obtained this way, the variety of the frames is extremely low, because the dataset is sourced from a limited number of videos.}
    \label{tab:dataset_comparison}
    \vspace{-0.3cm}
\end{table}

\section{Reproducibility and Open Resources}  
To foster reproducibility and facilitate further research in the field, we provide our code at \newline {\url{https://kiedani.github.io/WACV2026/}}.

\end{document}